# HIGH-RESOLUTION IMAGE RECONSTRUCTION WITH UNSUPERVISED DEEP LEARNING AND NOISY DATA APPLIED TO ION-BEAM DYNAMICS FOR PARTICLE ACCELERATORS

F.R. Osswald[1]*, M. Chahbaoui[2], X. Liang[3]

[1] *IPHC, CNRS Nuclei & Particules, Univ. of Strasbourg, 67000, Strasbourg, France*
[2] *University of Strasbourg, 67000, Strasbourg, France*
[3] *Sorbonne University, 75006 Paris, France*

**ABSTRACT**

Image reconstruction in the presence of severe degradation remains a challenging inverse problem, particularly in beam diagnostics for high-energy physics accelerators. As modern facilities demand precise detection of beam halo structures to control losses, traditional analysis tools have reached their performance limits. This work reviews existing image-processing techniques for data cleaning, contour extraction, and emittance reconstruction, and introduces a novel approach based on convolutional filtering and neural networks with optimized early-stopping strategies in order to control overfitting. Despite the absence of training datasets, the proposed unsupervised framework achieves robust denoising and high-fidelity reconstruction of beam emittance images under low signal-to-noise conditions. The method extends measurable amplitudes beyond seven standard deviations, enabling unprecedented halo resolution.

KEYWORDS: Computer vision; Inverse problems; Image denoising; Machine learning; Deep Convolutional Neural network; Beam physics; Transverse phase-space distributions.

## 1. INTRODUCTION

Image reconstruction degraded by factors such as aging, blurring, fragmentation, lighting defects, unadjusted optics, distant vision, or propagation through inhomogeneous media represents a longstanding class of inverse problems, already addressed in the earliest convolutional network approaches [1, 2]. Today, deep learning algorithms excel across diverse domains, including nuclear magnetic resonance (NMR) in medical imaging, species classification in the life sciences, biological microscopy, astronomical object identification, and topological monitoring in satellite imaging [3-5].

In the domain of high-energy particle colliders—large-scale infrastructures dedicated to fundamental and applied physics—beam diagnostics are critical to ensuring both performance and operational safety [6]. With continuously increasing beam intensity and power, and the expansion of installations, minimizing and controlling beam losses has become a priority. This objective is achieved by monitoring beam transmission through precise measurements of parameters such as position, profile, and phase-space distributions. Such measurements require the detection of electric charges over a dynamic range up to $10^6$ including extremely low halo intensities encountered during experiments with test beams, together with an effective suppression of background noise. Traditional analysis tools are no longer sufficient to meet the accuracy and efficiency requirements of modern and future accelerators and colliders [7]. In this paper, we present a review of existing image processing and analysis techniques applied to data cleaning, contour recognition, and emittance figure reconstruction, examined through the lens of promising artificial intelligence methods. Furthermore, we propose and investigate a novel image-based approach employing convolutional filters and neural networks to advance beam diagnostics.

---

\* Corresponding author: francis.osswald@iphc.cnrs.fr

## 2. RELATED WORK

### 2.1 Noise

Noise is an unwanted byproduct of image processing. It can severely degrade not only visual quality but also the performance of high-level computer vision tasks. Consequently, noise suppression has long been a central objective in the field of image denoising. Noise can be broadly defined as any event external to the object under observation that introduces spurious pixel intensities, thereby degrading the fidelity of the acquired image. In practice, noise manifests as random or structured variations in pixel values that obscure the true signal and reduce the accuracy of subsequent quantitative analyses. From a measurement theory perspective, noise represents an additive or multiplicative component superimposed on the desired signal, often modeled as a stochastic process with specific statistical properties. Its impact is particularly critical in scientific imaging, where the preservation of fine structural details and accurate intensity distributions is essential for reliable diagnostics. The challenge lies in distinguishing noise from genuine signal features, especially when the noise distribution is unknown or varies dynamically with experimental conditions. In the context of image acquisition for beam diagnostics and other high-precision applications, noise not only diminishes perceptual image quality but also propagates errors into higher-level tasks such as contour recognition, phase-space reconstruction, and emittance measurement. As a result, robust denoising strategies are indispensable to ensure that the reconstructed images faithfully represent the underlying physical phenomena.

One of the fundamental challenges lies in distinguishing noise from the true signal in the absence of prior knowledge of the noise distribution. This difficulty is exacerbated by variable experimental conditions and fluctuating background noise, which hinder the development of universally applicable denoising strategies [8]. Classical statistical methods are still widely employed across operational particle accelerator facilities to extract information from beam diagnostics data. However, these approaches encounter significant limitations when confronted with noisy signals, irregularly shaped beam cross-sections, non-elliptical or non-Gaussian phase-space distributions, non-uniform densities, or parasitic contaminating beams [9, 10]. More recent studies have highlighted the lack of flexibility in conventional filtering techniques, particularly in noise rejection under low signal-to-noise ratio (SNR) conditions and in preserving the edges of regions of interest (ROI) in images generated by digital analysis tools [11]. Such shortcomings lead to information loss and diagnostic inaccuracies, which in turn compromise the measurement of statistical emittance (RMS). The resulting mischaracterization of beam properties hampers the proper matching between machine acceptance and beam emittance, ultimately becoming a source of malfunction in large-scale accelerator installations. The most widely used approach for RMS emittance measurement relies on applying an arbitrary threshold to distinguish useful signals from background noise. However, the presence of noise or the absence of even minimal signals (less than $10^{-4}$ of the integral beam intensity) can introduce substantial bias, leading to errors in the RMS calculation that may exceed 100%. Conventional dark-field subtraction, which records noise in the absence of beam exposure to the sensor proves inadequate as the transmission of the beam through the scanner itself generates additional noise contributions. The inability to properly account for the beam halo, combined with limited measurement resolution—characterized by high intensities in the beam core and lower intensities at the edges within the halo transition zone—remains a recurring challenge [12-14]. In addition, the variability of background noise over time, across experiments, and depending on instrument location and installation complicates the definition of a single, reliable noise model.

Our emittance scanner [15], designed to measure particle distributions within the beam, produces high-resolution images with noise levels reaching 1 mV (equivalent to 1 nA in beam intensity). These images may contain negative values and exhibit a SNR below unity for the halo region. Noise sources in collider and accelerator environments are diverse, including parasitic signals inherent to the experimental setup, power supplies, electric motors, scanner electronics, cables, connectors, and technical equipment (lighting, ventilation, vacuum systems, etc.) [16]. The absence of a universal model and the lack of noise-free reference images (ground truth) necessitate the adoption of blind denoising strategies. To characterize background noise (BGN) specific to our scanner, several theoretical models were considered, including the signal-dependent noise level function (NLF), Poisson, Speckle, salt-and-pepper, additive white noise, Gaussian, Rician, statistical, random, median, quantum, impulse, and frequency-domain noise. Preprocessing tests were conducted using standard filtering techniques, such as convolution, Gaussian, Fourier transform-based (FFT), Kalman, and wavelet-based denoising. The most representative results will be presented later in the article.

### 2.2 Beam emittance and halo

The emittance of a beam in accelerator physics is a fundamental parameter that characterizes the distribution of particles within phase space and directly reflects the beam's quality and focusing properties [17]. In beam physics, transverse emittance is particularly important, as it quantifies the spread of particle positions $x$ and momenta $x'$ (or $y$ and $y'$) in the $xx'$ (or $yy'$) plane perpendicular to the beam trajectory. Accurate determination of this quantity requires the calculation of statistical measures that describe the beam's spatial and angular distributions, thereby providing insight into its confinement and transport efficiency. By analyzing the two-dimensional transverse emittance, one can evaluate the beam's suitability for applications demanding high precision, such as collider experiments or synchrotron light sources. This characterization thus serves as a critical step in optimizing beam dynamics and ensuring the stability of accelerator performance. In the case of two-dimensional transverse emittance ($xx'$ plane formulation taken in the following), the characterization requires the calculation of second-order statistical moments. Following quantities are defined for non-centered data.

Weighted means
$$\langle x \rangle = \frac{\sum_i I_i x_i}{\sum_i I_i}, \qquad \langle x' \rangle = \frac{\sum_i I_i x'_i}{\sum_i I_i} \qquad (1)$$

Second-order weighted moments
$$\langle x^2 \rangle = \frac{\sum_i I_i x_i^2}{\sum_i I_i}, \qquad \langle x'^2 \rangle = \frac{\sum_i I_i x'^2_i}{\sum_i I_i}, \qquad \langle xx' \rangle = \frac{\sum_i I_i xx'_i}{\sum_i I_i} \qquad (2)$$

variance and covariance
$$\sigma_x = \sqrt{\langle x^2 \rangle - \langle x \rangle^2}, \qquad \sigma_{x'} = \sqrt{\langle x'^2 \rangle - \langle x' \rangle^2}, \qquad \sigma_{xx'} = \sqrt{\langle xx' \rangle - \langle x \rangle \langle x' \rangle} \qquad (3)$$

With $I_i$ the intensity of each particle of the $x$ and $x'$ physical quantities. The root-mean-square (RMS) transverse emittance in one plane (horizontal) is formally defined as

$$\varepsilon_{RMS} = \sqrt{\sigma^2_x \sigma^2_{x'} - \sigma^2_{xx'}} \qquad (4)$$

where $x$ denotes the transverse position, $x'$ the angular divergence, and the brackets $\langle \rangle$ represent ensemble averages over the particle distribution. An analogous expression holds for the vertical plane, $\varepsilon_y$.

Twiss parameters
$$\alpha_{xx'} = -\frac{\sigma^2_{xx'}}{\varepsilon_{RMS}}, \qquad \beta_{xx'} = \frac{\sigma^2_x}{\varepsilon_{RMS}}, \qquad \gamma_{xx'} = \frac{\sigma^2_{x'}}{\varepsilon_{RMS}} \qquad (5)$$

The transverse beam envelope can be quantified by estimating the surface area associated with the RMS emittance at a specified standard deviation level, see Fig. 1. For a Gaussian distribution, the contour of constant probability density forms an ellipse in phase space. The area of the ellipses encompassing $N*RMS$ particles (or N sigma) is

$$A_N = \pi \cdot N^2 \cdot \varepsilon_{RMS} \qquad (6)$$

This relation highlights the direct proportionality between emittance and the geometric extent of the beam, providing a rigorous measure of beam confinement. Similarly, the following equivalence for N=1 holds: $A_{RMS} = \pi \cdot \varepsilon_{RMS}$.

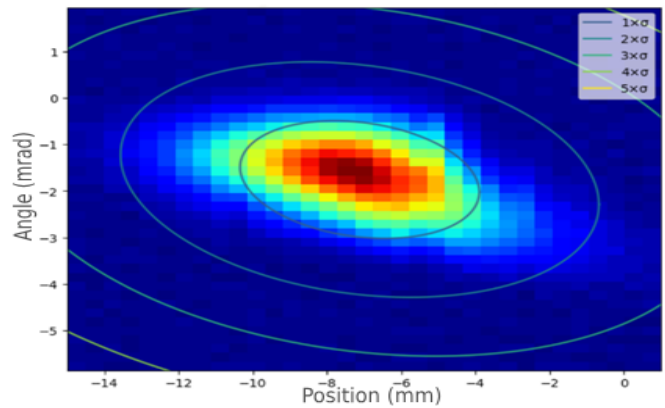

Figure 1: Representation of beam particles distribution in 2D transverse phase-space and associated RMS ellipses of emittance. Smallest ellipse in the center defines the one sigma reference.

Some images produce inaccurate ellipses for various reasons: incorrect scaling, pixel ordering issues, etc. and so need preprocessing or correction. The emittance ellipse is clearly affected by the noise present in the image, so it can be used as a visual and quantitative metric.

Beam halo refers to the low-density population of particles that exhibit large transverse oscillation amplitudes relative to the core of the beam. These particles can extend beyond the nominal beam envelope and impinge upon the aperture of the beamline enclosure. Such interactions give rise to uncontrolled beam losses, which in turn may induce radio activation of surrounding components and cause significant damage to accelerator structures. The characterization and mitigation of beam halo are therefore critical for ensuring operational stability, minimizing radiation hazards, and preserving the integrity of high-intensity accelerator facilities. For a Gaussian beam, the halo cross-section can extend to amplitudes as large as eight standard deviations from the beam core. Therefore, beam halo measurements require measurements of beam profiles ($x$, $y$ positions) with a very high dynamic range, at least $10^6$. Remarkably, fewer than 10% of the particles may occupy regions that correspond to more than 70% of the total transverse beam area. This disproportionate spatial distribution highlights the critical role of halo particles in determining beam–aperture interactions, as a relatively small fraction of the beam population can dominate losses and activation processes.

## 2.3 Convolutional filters and usual machine learning algorithms

The initial stage of the investigation focused on simple algorithms, including convolution-based filters and basic machine learning (ML) models for binary classification and clustering. These approaches were applied to distinguish between useful signals and noise, effectively implementing a "good/no-good" decision framework equivalent to signal versus background discrimination. The application-driven image restoration network should be capable of removing noise in images and preserving their semantic aware details. The main characteristics of the required algorithm include robustness of cleaning/denoising across a wide range of conditions without the need for manual parameter tuning, applicability to diverse noisy images and artifacts, and stable performance under varying noise levels. Certain difficulties should be overcome, such as loss of fine details, aberrations in the halo region, poor contour definition leading to weak beam characterization and transport performance, and reduced denoising efficiency at low signal-to-noise ratios [18].

Among classical approaches, the median filter remains one of the most widely used techniques. This non-linear method replaces each pixel with the median value of its neighborhood and, when properly designed, is effective at preserving image details. Traditional statistical denoising and cleaning methods are often poorly adapted to the specific features encountered in beam physics. They typically rely on assumptions about scanned images or noise models, and in most cases require user intervention and non-trivial parameter tuning to accommodate experimental variability. For example, many conventional algorithms restore images by minimizing the mean squared error (MSE), a criterion that is misaligned with perceptual image quality and prone to over-smoothing textures, thereby erasing fine structural details [19]. In contrast, modern ML—particularly convolutional neural networks (CNNs)—offer greater robustness and adaptability without user intervention. Unlike classical filters, CNNs can learn complex noise distributions directly from data, eliminating the need for manual parameter tuning. They are capable of preserving high-frequency details such as edges and textures, while simultaneously rejecting noise across a wide range of signal-to-noise ratios. Advanced methods often perform a global analysis of point clouds, producing outputs that depend on the entire dataset, which is necessary for extracting global properties such as semantic classification [19-26]. Furthermore, ML-based approaches can generalize to diverse experimental conditions, including irregular beam shapes, non-Gaussian phase-space distributions, and parasitic signals, where classical statistical methods often fail. This shift from hand-crafted filters to data-driven models reflects a broader trend in image restoration research. While classical methods remain valuable for their simplicity and interpretability, AI-based approaches provide the flexibility and perceptual alignment necessary to meet the accuracy and efficiency demands of modern accelerator diagnostics [27].

In general, the objective of image restoration (IR) is to recover the latent clean image $x$ from its degraded observation $y = Hx + w$, where $H$ represents the degradation operator and $w$ denotes additive white Gaussian noise defined by a standard deviation [28]. By specifying different degradation matrices, distinct IR tasks can be defined:

- Image denoising, when H is the identity operator;
- Image deblurring, when H is a blurring operator;
- Image super-resolution, when H is a composite operator combining blurring and down-sampling.

Density-Based Clustering like DBSCAN can be used for clustering 2D point clouds by grouping points based on density into binary foreground/background segments. For the recognition and detection of point-cloud edges under high-noise conditions, several CNN architectures have demonstrated notable efficiency. PointNet and its variant ++ which consider hierarchical structures to capture fine geometric details, whereas PointCNN introduces adaptive convolution operations tailored to irregular point distributions. KPConv (Kernel Point Convolution) applies low-level point-based convolutions, DGCNN (Dynamic Graph CNN), leverages dynamic graph structures to model relationships among points. Finally, PointSIFT integrates both local and global features, enhancing edge detection and structural characterization in complex point-cloud data. The intrinsic nature of particle distributions within a beam tends to blur boundaries, as the distribution extends indefinitely with decreasing density and lacks sharply defined contours. Despite this challenge, classification and clustering algorithms will be evaluated to enhance image analysis. Their application is expected to provide guidance for the development and selection of more advanced modeling approaches.

## 2.4 Deep convolutional neural networks

Deep learning models have reached a level of maturity that enables robust performance across diverse tasks in a large field of applications including computer vision, natural language processing, healthcare diagnostics, and autonomous systems. In computer vision for example, it has significantly reduced the error rate in object recognition, with an exponential decrease since 2010 [29]. The principal constraints of this study—namely the absence of ground-truth reference images, limited availability of training datasets, lack of a well-defined noise model, variability in target shapes, and low SNR—motivated the exploration of advanced unsupervised learning approaches. In particular, unsupervised deep convolutional neural networks (DCNNs), diffusion-based models, and Transformer architectures were investigated as promising candidates to overcome these limitations and provide robust solutions for image restoration and beam diagnostics. Following a comprehensive survey of image restoration architectures presented in the literature, we selected the Deep Image Prior (DIP) model as the foundation of our approach [30]. DIP is distinguished by its ability to perform image denoising without reliance on external training datasets, instead exploiting the inherent statistical structure encoded within the CNN architecture itself. By optimizing the network parameters directly on the corrupted image, DIP leverages the natural image priors implicit in the network design to progressively recover clean visual content. This self-supervised mechanism eliminates the need for labeled data or large datasets, which is often scarce or unavailable in many practical scenarios, thereby making DIP particularly advantageous for unsupervised learning tasks. Moreover, its capacity to generalize across diverse noise distributions without explicit pre-training and noise specification underscores its suitability for applications where adaptability and data efficiency are critical. DIP is a convolutional network featuring an encoder–decoder architecture of the U-Net type (so called hour glass with skip connections). The central idea is that the structure of this deep convolutional neural network (DCNN) inherently favors natural image statistics (or resulting from a physical process) during optimization. These so-called priors arise from the capture of patches, patterns, features, and pixels exhibiting similar characteristics across multiple scales. Such priors encode structural redundancies in the data (also called self-similarities) and provide constraints that guide the learning process toward more robust and generalizable representations. When the weights are adjusted to reconstruct a noisy image, the network first captures the underlying clean structure before fitting the noise. The selective impedance between signal and noise cannot be attributed to an innate ability comparable to the mechanisms by which humans perceive and interpret the external world [31], but reflects an intrinsic architectural bias (some authors have suggested the existence of a resonance between the structural properties of the data and those of the model). As a result, DIP enables image denoising without requiring any external dataset for training, clean (ground truth)/noisy pairs, annotations, or explicit prior models. It is particularly recognized for its ability to preserve fine details and local textures in the absence of supervision. The main limitation of DIP lies in its sensitivity to overfitting when the stopping criterion and hyperparameters are not properly defined. Since optimization is performed on an image-by-image basis, the process can be computationally expensive, and early stopping (ES) is critical to prevent adaptation to noise. The model is highly dependent on hyperparameter choices and architectural design, and it does not scale efficiently to large image datasets. These issues will be addressed in the following by proposing solutions tailored to images produced by scanners used in particle accelerators to measure beam distributions. For example, in order to mitigate overfitting, DIP has been combined with self-supervision strategies, incorporating automated stopping mechanisms through regularization and perceptual or physics-informed criterion specific to the application. DIP is unsupervised, thus standard techniques like Correlation-Driven Stopping Criterion (CDSC) can't be used directly.

The inferred image $x$ is obtained as the output of the model $f_\theta(z)$, where $z$ denotes a random input tensor and $\theta$ represents the set of trainable parameters. Formally,

$$x = f_\theta(z) \qquad (7)$$

This formulation highlights the generative nature of the model, in which the random tensor $z$ is mapped through the parameterized function $f_\theta$ to produce the reconstructed image $x$ from a corrupted version $x_0$. With a regularizer $R(z)$ of the model, inferred image $x^*$ results from an energy minimization problem and becomes

$$x^* = \mathrm{argmin}_x\, E(x, x_0) + R(x) \qquad (8)$$

By injecting stochastic perturbations into the input space as a random input tensor at the first and alternatively each iteration during the training of the model, $z$ prevents overfitting and promotes smoother optimization trajectories. The network parameters are updated to minimize the discrepancy between the corrupted input and the reconstructed output, thereby progressively suppressing noise while preserving structural details. This repetition continues until the optimization reaches a stable state, where further iterations yield negligible improvements in image quality. Such convergence behavior reflects the capacity of the model to exploit the implicit priors encoded in its architecture, ultimately producing a restored image that balances fidelity to the original content with effective noise reduction, see Fig. 2.

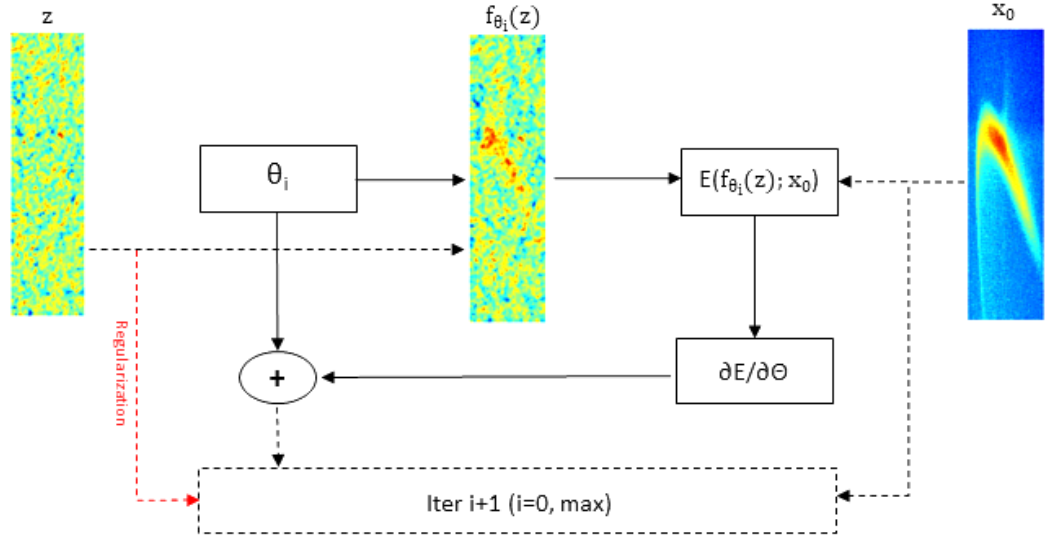

Figure 2: Flow diagram of the image denoising process, adapted from [30]. The weight parameters $\theta$ are randomly initialized and iteratively updated to minimize the cost $argmin_\theta\ E(f_\theta(z); x_0)$ between the noisy input $x_0$ and the restored image $f_\theta(z)$. The variable $z$ denotes an arbitrary noise tensor introduced to initialize the training and regularize the convergence of the neural network $f$. The iterative optimization loop quoted *Iter i+1* is executed repeatedly until the restoration process converges to a visually coherent image.

Other approaches, such as Noise2Noise, Denoiseg and Noise2Self [32, 33], employ self-supervision by masking target pixels and learning to predict them from the surrounding noisy context, under the assumption that noise is independent of the neighborhood used for prediction. According to NTIRE and CVPR challenges, these methods are more effective when a large set of noisy images from the same domain, though not necessarily from the exact application domain, is available. Like DIP, they do not require clean/noisy pairs. Standard training on a complete dataset enables fast inference and stable performance under independent additive noise, although performance may degrade for more complex noise patterns that are spatially correlated with the signal or when masking violates the independence assumption. Careful training is nevertheless required to avoid blurring of fine details.

## 2.5 Other options

Compared to generative adversarial networks (GANs), diffusion-based models [34-37] and transformers [38, 39], which often demand extensive pre-training and computational resources, DIP offers a lightweight yet effective alternative that generalizes across diverse noise distributions. Its ability to adapt to individual images without prior knowledge underscores its relevance for unsupervised learning scenarios, making it a compelling choice for our application.

## 3. METHODOLOGY

Prior to training the network, a thorough data analysis of the 2D phase-space distributions was carried out on the emittance figures to ensure the reliability of the learning process. Such analysis (with tools known as EDA, exploratory data analysis) enables the identification of corrupted samples, class imbalance, and systematic noise that could otherwise bias the model. This step also informs the selection of hyperparameters and training strategies, thereby improving efficiency and reducing the risk of overfitting. Ultimately, data analysis enhances transparency and reproducibility, ensuring that the network is trained on representative and high-quality inputs.

The first step was to collect a large number of images representing the beam emittance in order to create the most diverse selection possible, with typical and atypical cases, irregular shapes, structures with defects, and images with varying degrees of noise. From the initial collection originating from six different installations, therefore with six different beams, experimental setups, emittance scanner settings, measurement procedures, thus a variety of phase-space distributions, we counted 2000 noisy images. Among them, a large number were unusable (image without beam, bad settings, poor resolution, off-center beam, and were automatically discarded by a sorting program to ultimately retain 10% of them. By selecting a subset of representative images from the several facilities, we aimed to characterize their distinctive features. To this end, a series of standard data analyses were conducted, including basic statistical evaluation, assessment of noise and signal distributions, analysis of beam intensity variations, correlation and distribution analyses as Fast Fourier Transform (FFT). As preprocessing, a comparative analysis was conducted across the datasets against several theoretical distributions, including Gaussian, Uniform, Poisson, Speckle, Rician, Salt-and-Pepper, Exponential, and Gamma. Due to the absence of obvious frequency components in the noise distribution (non-stationary physics), FFT type analysis did not provide conclusive results and appears unsuitable. This has led us to consider clustering approaches such as DBSCAN and related algorithms. In order to evaluate the fundamental behavior of the restoration models, to better understand, optimize their performances, and realize a selection of the most efficient ones, the denoising process has been carried out with various quantitative metrics (PSNR, TV, physics-informed metrics, etc.) on a large number of images.

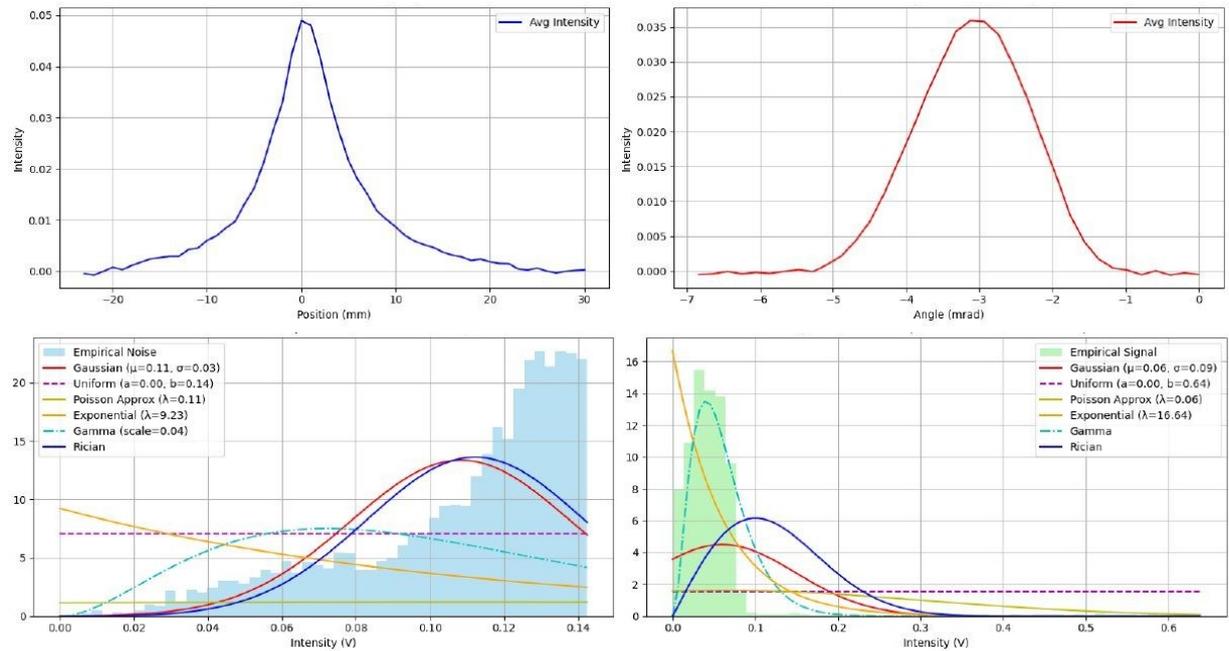

Figure 3: Data analysis of a standard beam profile showing radial distributions of particle positions and divergence angles (top left and right). Corresponding intensity distributions of the beam particle positions and angles and comparison with a selection of theoretical noise distributions (bottom left and right). The parameters of the theoretical noise were fitted in order to define optimized values and data shifts were applied to enable comparisons.

Basic statistics carried out on the signal and noise data consist in number of pixels evaluation, image dimension definition (width × height), and intensity analysis: maximum, minimum and mean intensity, and standard deviation. In signal processing and physical measurements (e.g. intensity, voltage), negative values may arise due to electronic noise, baseline drift, subtraction artifacts, and related effects. Many theoretical distributions, however, require non-negative values. Since our raw noise data contains negative entries, direct comparison would be invalid without applying a shift. Therefore, we first shift the noise and signal regions to enable shape similarity testing. Shifting serves as a practical workaround to map the data into a valid domain for visualization and fitting. Importantly, this operation preserves the distributional shape and modifies only its location along the axis, making it suitable for shape-based analyses such as modeling and denoising. The exact shift amount must be recorded to ensure accurate image reconstruction. Nevertheless, shifting should be avoided when the zero value carries physical meaning (e.g. voltage measurements) or when comparisons are made to distributions that inherently allow negative values (e.g., Gaussian, Uniform). In the second stage of processing, the signal was separated from noise using a predefined intensity threshold in order to study their characteristics separately, see Fig. 3. Pixels with values greater than or equal to the threshold were classified as signal, whereas those with values below the threshold were designated as noise. This binary partitioning provided a systematic criterion for distinguishing meaningful image content from background fluctuations. The distributions of particle positions and angles did not conform to any known theoretical model. However, when the threshold was increased, the resulting distribution tended to approximate a Gaussian. No correlation was detected among the noise components. Then, due to significant variation in image properties (e.g., size, beam shape, noise distribution, scale, resolution, ROI centering, etc.), dataset-wide normalization was not feasible reinforcing our choice of DIP for per-image denoising.

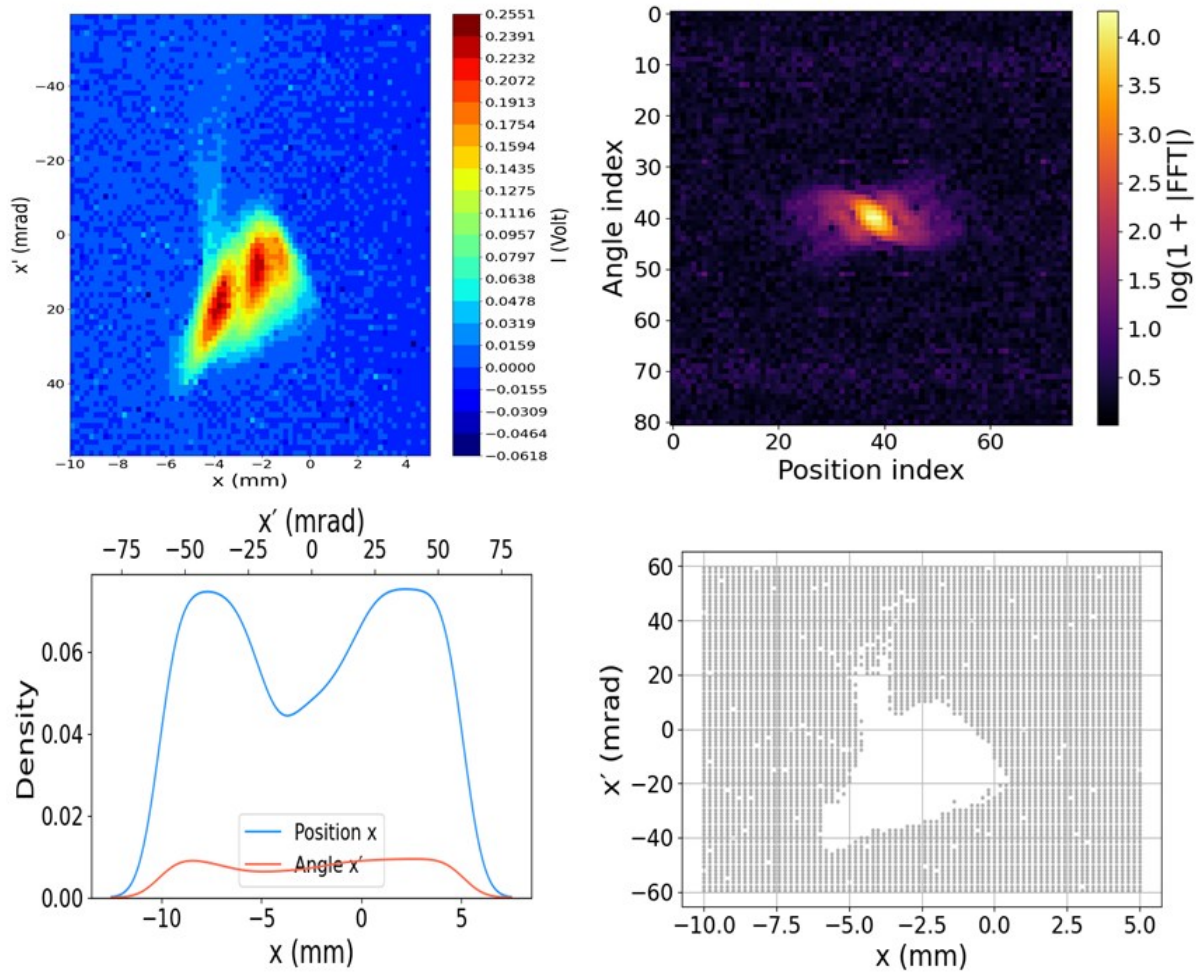

Figure 4: High-resolution image obtained with a beam scanner and showing 2D transverse phase-space distribution of an ion-beam accelerated at a few kV (top left), corresponding 2D-FFT analysis of pixelized phase-space distribution (top right), radial distribution of the particle positions and divergence angles (bottom left), and image segmentation obtained with DBSCAN to discriminate background noise (correction factor equals -0.009).

Other analyzing tools were tested in order to discriminate the signal from the noise, and extend the data analysis. Among them are the binary classification and clustering algorithms for 2D point clouds DBSCAN, HDBSAN, and GMM [40-43]. The DBSCAN algorithm (Density-Based Spatial Clustering of Applications with Noise) is primarily grounded in concepts from geometry and graph theory rather than topology for the identification of clusters. Each data point is treated as a node in a graph, and edges are established between nodes when their pairwise distance falls below a predefined threshold. Clustering then corresponds to the identification of connected components within this graph, see Fig. 4. Geometry plays a central role in defining distances between points and their neighborhoods. Euclidean distance (or alternatively other distance metrics) is typically employed to determine whether two points are neighbors. Clusters emerge as regions of high point density, a process inherently based on geometric considerations. Although DBSCAN does not explicitly rely on topological concepts, indirect connections exist: the notions of neighborhood and dense connectivity are loosely related to continuity and neighborhood structures in topology. Recent studies have explored the use of algebraic topology to enhance or better understand clustering methods, but such approaches are not foundational to DBSCAN. In summary, DBSCAN is primarily anchored in geometry and graph theory. The algorithm seeks to identify dense regions of points. It begins by selecting an arbitrary point; if this point has a sufficient number of neighbors (at least *MinPts*), it initiates a cluster. A cluster is defined as a set of densely connected points. The process is iteratively expanded by including all neighbors within the radius Epsilon (*Eps*), until no further points can be added. A point is considered part of a cluster if it has at least *MinPts* neighbors within the radius *Eps*; otherwise, it is classified as noise. The parameter *Eps* is critical, as it defines the neighborhood radius within which points must lie to be considered neighbors. This enables DBSCAN to detect clusters of arbitrary shape and size, based solely on point density. A straightforward clustering algorithm such as DBSCAN can classify pixels effectively; however, tuning its parameters for each image is both time-consuming and computationally demanding. The results remain unsatisfactory because parameter optimization is incomplete, and no ground truth is available to serve as a reference for evaluation or parameter fitting. Ideally, background noise should exhibit a lower density than the ROI. Nevertheless, the intermediate region occupied by the halo is poorly resolved, preventing a clear binary classification.

Hierarchical DBSCAN (HDSCAN) is an extension of DBSCAN that builds a hierarchy of clusters. It picks the most stable clusters from that hierarchy and adapts to varying densities, handling both beam core and halo properly, see Fig. 5. Finally, its advantage is that there is only one parameter to tune (no need to tune *Eps*).

The Gaussian Mixture Model (GMM) was evaluated as an alternative clustering approach. GMM assumes that the data are generated from a mixture of several Gaussian (normal) distributions, with each cluster represented as a Gaussian component in the feature space. This method employs probabilistic soft clustering, whereby each point is assigned to all clusters with a certain probability. In our context, we hypothesize that the beam and its surrounding halo form distinct Gaussian-like groups, which can be represented by interpretable ellipses defined by the cluster means and covariance matrices. Among the advantages, there are no parameters to tune, and the number of clusters which can be set to 2 or 3 for our study. Since GMM assumes a Gaussian distribution of the beam, it may detect the signal more effectively in some cases where the underlying distribution is close to Gaussian. Certain limitations emerged when the data exhibited non-Gaussian distributions or non-convex cluster shapes, and considering S/N rejection, higher performances are expected with frameworks based on deep learning.

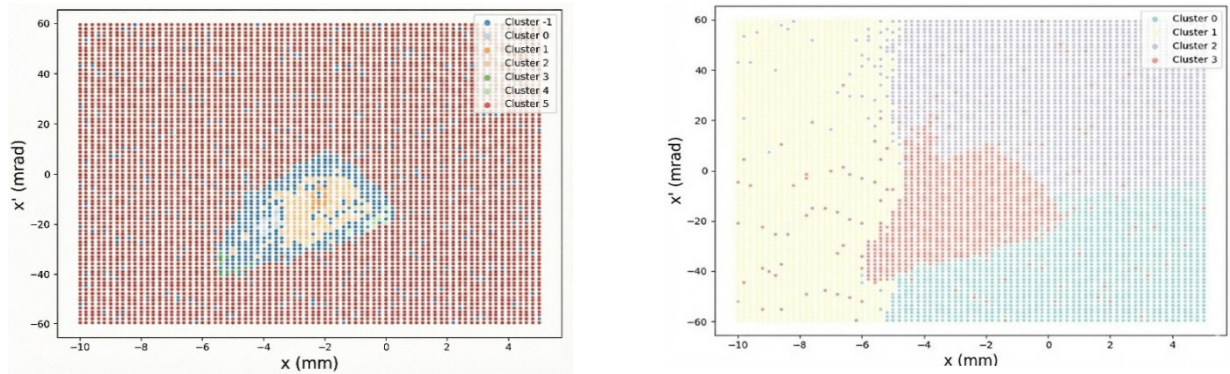

Figure 5: Similar high-resolution image segmentation to discriminate background noise and identify region of interest (beam core) obtained with HDBSCAN (left), and GMM point cloud clustering models (right).

Data files of the grey scale images are converted from original scan format (.dat) to .csv to facilitate analysis in Python with Pandas in association with other libraries, see annex A. The digitized images are archived as standardized files that include the scanner settings, such as spatial resolution and pixel intensity values. Each pixel of an image represents the intensity measurement of a beam sample taken during one scan step. The number of steps defines the resolution of the instrument. For deep learning models, images are often processed as float32 with values between 0.0 and 1.0. To save or display them correctly with classic libraries (OpenCV, PIL, etc.), they are converted back to the standard uint8 format (unsigned integer 8 bits), i.e. integers between 0 and 255. This is the typical format for RGB or greyscale images. Each two-dimensional image in grayscale is subsequently represented as a rank-3 tensor, where the two first dimensions correspond to the beam particle positions (mm) and their momentum (angular divergence in mrad), and third dimension corresponds to signal intensity (converted in mV). The processing of tensor rather than image led to substantial computing time reduction, by a factor 2 in most of the cases. Formally, an image can be expressed as $I \in \mathbb{R}^{m \times n \times 1}$, where m and n denote the number of rows and columns, respectively, and the singleton dimension corresponds to the grayscale intensity channel. The tensor dimensions range from 10,000 to 250,000 corresponding to the total number of pixels in the image (rows × columns). It depends on the pixelized image resolution (12 bits) and the spatial extent of the beam region being scanned. We then proceed to normalization of the data with the numpy library to ensure that input features are on comparable scales, which improves training stability, speeds up convergence, and prevents features with larger ranges from dominating the learning process.

The standard computer configuration used during the developments was: laptop intel i5 64 bits, 2.7 GHz, RAM 16 GB, 1 GPU NV RTX 3050 with 6 GB virtual RAM (power consumption ~70 W). The typical computation time remains below four minutes for $10^5$ pixels and a few hundred iterations, corresponding to an average cost of well under one second per iteration. This performance level enables rapid parameter sweeps and makes the approach suitable for routine use in large-scale or iterative optimization studies.

## 4. MODEL FOUNDATIONS

### 4.1 DIP instantiation

The model employed in this study is based on the skip architecture, a convolutional encoder–decoder network with skip connections, conceptually similar to U-Net [30]. This design enables the reuse of low-level features from earlier layers during decoding, thereby preserving spatial structure and improving reconstruction fidelity. The principal architectural and parameter settings are summarized below.

Architecture: convolutional encoder–decoder (32 filters per layer) with skip connections, facilitating the transfer of fine-grained spatial information across layers.

Input channels (1): the network processes single-channel input data, such as grayscale images or beam intensity maps.

Output channels (1): the output is likewise a single-channel image, corresponding to the reconstructed beam distribution.

Number of scales (2): the network operates across two down sampling/up sampling scales, encoding the input into coarser representations and reconstructing it twice to refine details.

Channel configuration:
skip_n33d = 32: 32 filters per down sampling layer in the encoder path.
skip_n33u = 32: 32 filters per up sampling layer in the decoder path.
skip_n11 = 2: 2 filters allocated to each skip connection branch.

Padding mode: reflection padding is applied to mitigate edge artifacts during convolution.
Up sampling mode: bilinear interpolation is used for up sampling feature maps in the decoder, ensuring smooth reconstruction.

Activation function: a LeakyReLU activation is applied after each convolutional layer to introduce non-linearity and enhance representational capacity.

Final activation: no sigmoid activation is applied at the output layer (need_sigmoid = False), this is consistent with the regression nature of the task (like beam intensity profile and emittance figure reconstruction).

### 4.2 Loss function

The total loss used to train the DCNN model is a custom loss function with weighted combination of four components:

**Weighted Mean Squared Error (MSE).** This term penalizes the squared difference between the network output and the input image, giving higher importance to high-intensity regions using a weight map (the center of the beam is more important than the edges of the image):

$$L_{\text{MSE}} = 1/N \sum_{i,j} w_{i,j} (\hat{I}_{i,j} - I_{i,j})^2 \qquad (9)$$

The weights $w_{i,j}$ typically emphasize bright areas or regions with high variance.

**Mean Absolute Error (MAE).** This is the $L_1$ loss between the network output and the input image:

$$L_{\text{MAE}} = 1/N \sum_{i,j} |\hat{I}_{i,j} - I_{i,j}| \qquad (10)$$

It is less sensitive to outliers than MSE and encourages sharper reconstructions, see Fig. 6.

**Total Variation (TV).** This regularization term penalizes spatial variations in the output image:

$$L_{\text{TV}} = \sum_{i,j} |\hat{I}_{i+1,j} - \hat{I}_{i,j}| + |\hat{I}_{i,j+1} - \hat{I}_{i,j}| \tag{11}$$

It promotes piecewise smoothness and reduces spurious noise without severely blurring edges.

**Gradient Difference Loss (GDL).** This term aligns the spatial derivatives (gradients) of the output and input images:

$$L_{\text{GDL}} = \sum_{i,j} |\nabla_x \hat{I}_{i,j} - \nabla_x I_{i,j}| + |\nabla_y \hat{I}_{i,j} - \nabla_y I_{i,j}| \tag{12}$$

It ensures that the edge structure and texture are preserved in the output.

**Total Loss Function.** The full loss is a linear combination of the above terms:

$$L_{\text{total}} = w_{\text{mse}} \cdot L_{\text{MSE}} + w_{\text{mae}} \cdot L_{\text{MAE}} + w_{\text{tv}} \cdot L_{\text{TV}} + w_{\text{gd}} \cdot L_{\text{GDL}} \tag{13}$$

By manually tuning the weights ($w_{\text{mse}}, w_{\text{mae}}, w_{\text{tv}}, w_{\text{gd}}$), we control the balance between fidelity to the input and desirable image properties like smoothness, sharpness, and structural integrity.

Table 1: Description of the weights and tuning conditions

| Loss term | Description | Use condition |
|---|---|---|
| w_mse | Penalizes large squared differences; focuses on preserving intensities | Use when the output has intensity distortions or noise |
| w_mae | Penalizes absolute differences; less sensitive to outliers than MSE | Use when the output is blurry or contains sharp spikes |
| w_tv | Total Variation, encourages spatial smoothness | Use when the output has speckle noise or unnatural texture |
| w_gd | Gradient Difference, preserves gradient/texture/edges | Use when you want to maintain fine edge structures and avoid over-smoot |

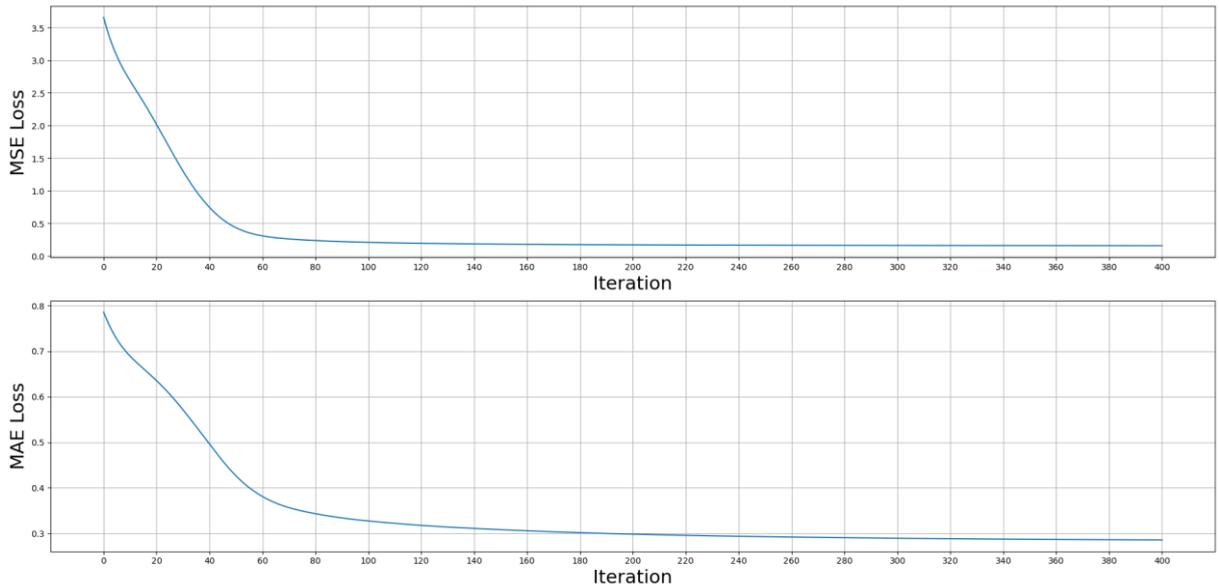

Figure 6: MSE and MAE loss functions show convergence after less than 380 iterations.

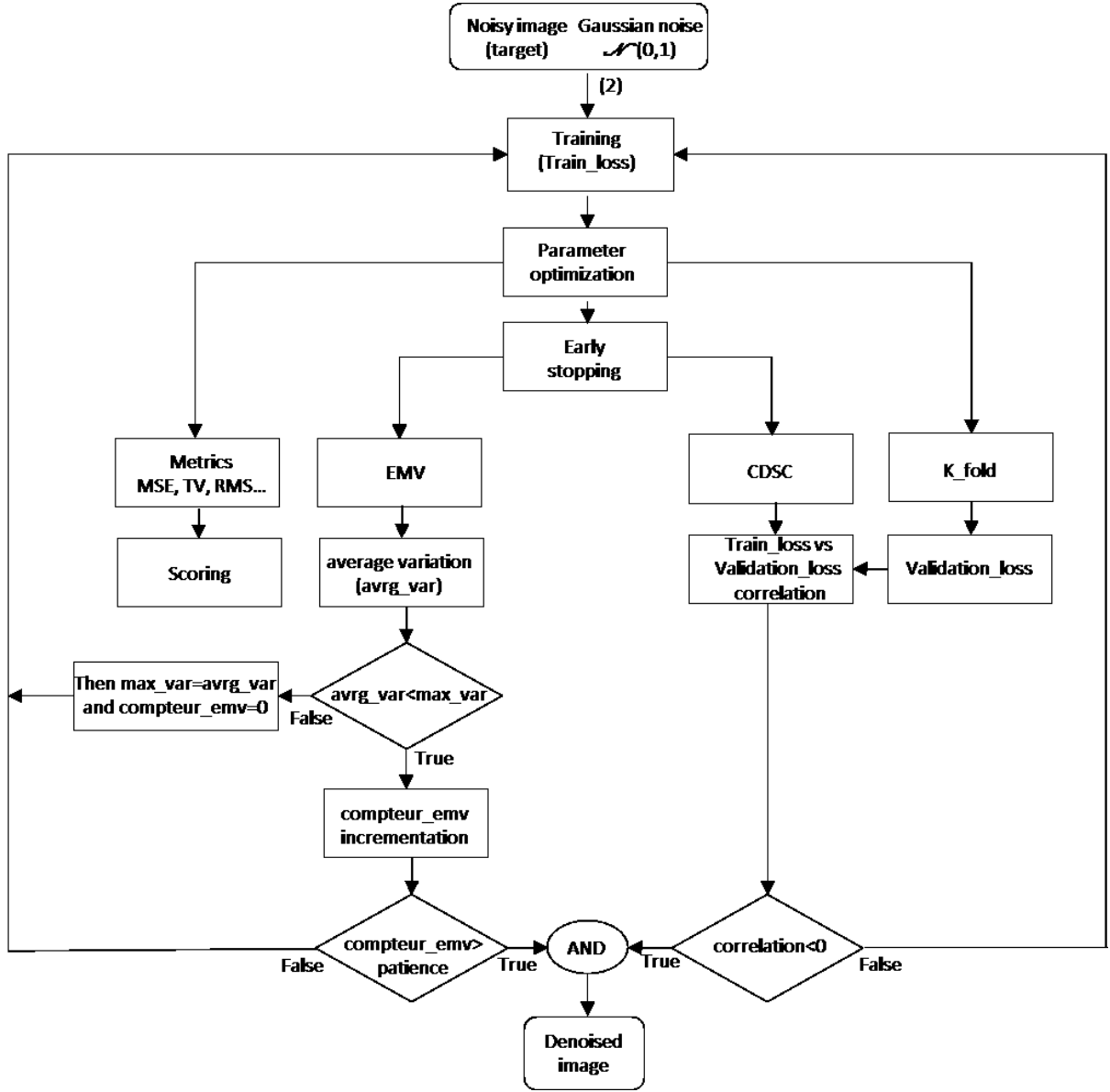

Figure 7: Logic flow diagram of the DCNN model performing the denoising of the HR images with early stopping strategy. The training loss is computed with masked outpouts, the average variation is computed on a sliding window. Thus, not all pixels from the original noisy image are used to optimize the model, only the remaining pixels after masking the validation set used for early stopping are considered during training. The « patience » parameter defines the maximum number of iterations the loop can tolerate without improvement before stopping. Improvement means « compteur_env » remains below the patience threshold. The « max_var » variable is set initially to « -inf » and keeps changing during the iterations.

Adding a small amount of noise to the input data at each iteration during training is an optional regularization technique. Its main purpose is to stabilize the optimization process and prevent the network from overfitting. This helps prevent the model from replicating input noise or other artifacts of the corrupted input image. The noise acts as a form of stochastic regularization similar to the dropout function. If the parameter defining the amount of noise (reg_noise_std) is too large, the network may fail to converge due to too much instability. If it is too small, the regularization effect diminishes. Several theoretical noise models were systematically evaluated to assess their effectiveness within our framework, and the experiments demonstrated that Gaussian noise yielded the most favorable image restoration performance. It turned out that normal or Gaussian noise with a mean value of 0 and a standard deviation of 1 yielded good results contributing to stabilization of gradient descent during training. Nevertheless, the impact of noise regularization is highly situational and uniform noise can create outliers that might mislead the optimization process. Its effect is often minimal when the network's architecture already acts as a powerful prior for the given data. Consequently, while the technique can be a useful fine-tuning tool in some scenarios, its contribution may be negligible in others.

## 4.2 Evaluation metrics

The model was deployed with random Gaussian noise as input and the noisy beam image as the reconstruction target. We evaluated and monitored the denoising results using both perceptual and quantitative metrics: Total Variation (TV), Gradient Descent Loss (GDL), NIQE, BRISQUE, Laplacian Var, Tenengrad, beam area (RMS emittance), beam profiles, Pseudo Validation Loss, etc.
First category of evaluation metrics used during training and testing, consists of no-reference image quality metrics. These are designed to assess image quality without access to any ground truth.
- NIQE (Natural Image Quality Evaluator): compares statistical features of the current image to those learned from a database of natural (pristine) images, lower scores indicate more "natural" and likely better-quality images. Use for ES: it should initially decrease as noise is removed. Early stop when NIQE begins to increase or flatten, indicating potential overfitting to noise.
- PIQE (Perception-based Image Quality Evaluator): segments the image into patches and evaluates each for blockiness, blur, and noise, lower is better. Use for ES: early stop when PIQE stops improving or begins to deteriorate.
- BRISQUE (Blind/Referenceless Image Spatial Quality Evaluator): measures "naturalness" based on spatial statistics like local luminance normalization, lower score means better perceived quality. Use for ES: monitor BRISQUE score as the image gets denoised. Watch for stagnation or increases, which may indicate the network is fitting noise.

Second category includes statistical and gradient-based metrics, and complement previous mentioned metrics. These are not true no-reference Image Quality Assessment (IQA) metrics but are strongly correlated with noise, sharpness, and structure, making them useful for ES.
- Shannon Image Entropy: measures the amount of information or randomness in an image; noisy images typically have higher entropy. Use for ES: entropy should decrease as noise is removed. If entropy starts to increase again, the model may be reconstructing noise and early stop is recommended.
- Laplacian Variance (focus measure): computes the variance of the Laplacian of the image. It is sensitive to edge content; high variance indicates sharper images. Use for ES: Laplacian variance should increase initially. Stop training when it peaks and starts to drop (likely due to fitting noise).
- Tenengrad (Gradient-based focus measure): calculates image gradients (usually via Sobel operator) and sums squared magnitudes; high values indicate greater detail and sharpness. Use for ES: similar to Laplacian variance. Stop when Tenengrad peaks, before noise causes it to rise artificially.

Next category of evaluation metrics concerns the validation metrics performed for early stopping after initial training of the model. Pseudo-validation strategies are used to create an artificial or self-supervised validation signal.
- Generation of a random binary mask: randomly masks a small percentage of pixels during training (typically 5-10%). The training loss consists of the full reconstruction loss computed only on unmasked pixels (all observables). The validation loss is the error calculated exclusively on masked pixels. This feature allows to monitor prediction performance on masked pixels, and ends training when performance on masked pixels stops improving. It should be noted that if PIQE is evaluated on partial or masked images, the metric can fail or become meaningless due to small unmasked areas, high-frequency reconstruction noise, or PIQE's sensitivity to edges and contrast. Therefore, PIQE iss not implemented in final configuration.
- Generation of a K-Fold mask: divides input (image) into K distinct parts. The training process consists of three consecutive steps, namely the training on the K-1 parts, the validation on the remaining part, and the rotation through all K possible combinations. This method allows to track the average loss across all folds as a pseudo-validation score, and to compute the metrics separately for each fold. The process is stopped when the average validation error across folds plateaus (stopping criterion). It should be noted that when using the K-Fold pseudo validation mask, this strategy is integrated into the original training loop of the DCNN, see logic flow diagram in Fig. 7. Before entering the loop, we divided all image pixels into $K$ subsets ($i = 1, ..., k$). For each fold $i$, the pixels in the i-th subset were used as the validation set, while the model was trained and optimized using the remaining pixels. This process was repeated for each fold to ensure comprehensive validation coverage.

The results are significant. Denoising is efficient after 20-30 iterations, see Fig. 8. The denoising process is not time consuming, it rarely takes more than 30 minutes.

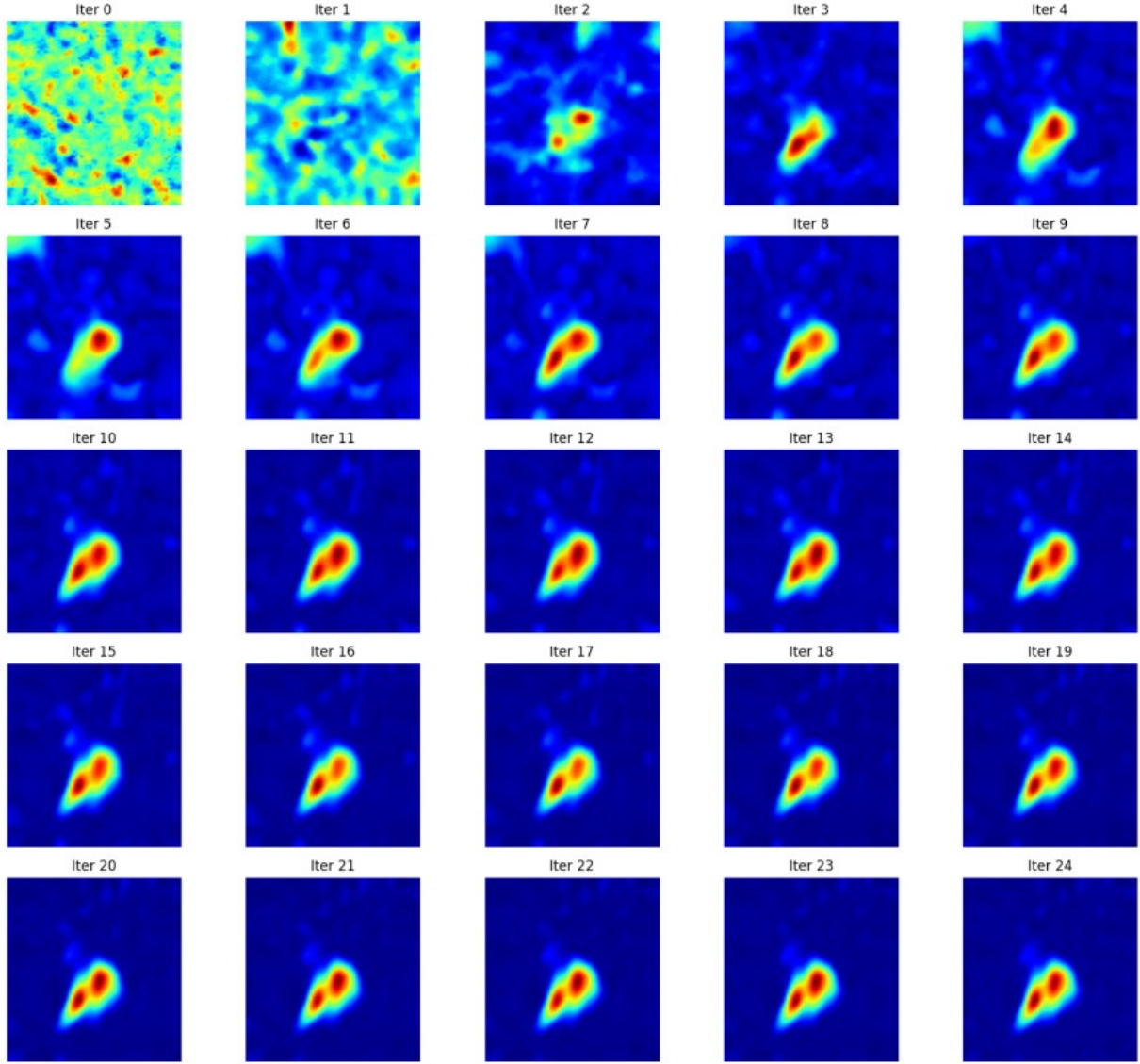

Figure 8: Reconstruction of an image representing the two-dimensional phase space distribution of an ion beam. The denoising process is clearly apparent after 20 iterations.

## 5. OPTIMIZATION

Model optimizing follows the procedure employed in DCNNs, namely the minimization of a loss function using stochastic gradient descent (SGD). Gradients of the loss with respect to the model parameters were computed via backpropagation, and the weights were updated iteratively according to the gradient descent rule. A fixed learning rate was applied during training after several fittings around 0.01, and convergence was monitored through the evolution of the validation metrics. To enhance stability and generalization, mini-batch training, appropriate weight initialization, and regularization techniques (e.g., early stopping) were employed.

Formally, the parameter update at iteration t is given by

$$\theta_{t+1} = \theta_t - \eta \nabla_\theta L(\theta_t) \quad (14)$$

where $\theta_t$ denotes the model parameters at iteration $t$, $\eta$ is the learning rate, and $\nabla_\theta L(\theta_t)$ represents the gradient of the loss function $L$ with respect to $\theta_t$.

During the optimization process, multiple loss functions and evaluation metrics were monitored to assess convergence and reconstruction quality. The Mean Squared Error (MSE) was first considered, as it quantifies the pixel-level deviation between the predicted image and the noisy input. A steady decrease in MSE indicates successful minimization of pixel-wise error by the network. However, while a lower MSE implies numerical closeness to the target image, it does not necessarily guarantee perceptual similarity.

The Peak Signal-to-Noise Ratio (PSNR), implemented via the *skimage.metrics* library, was also tracked. PSNR measures the ratio between signal and noise, with higher values corresponding to improved fidelity and reduced noise. Since PSNR is derived from MSE on a logarithmic scale, it aligns more closely with human visual perception, making it a complementary metric for image quality assessment.

In addition, the Mean Absolute Error (MAE) was employed. Unlike MSE, which squares deviations and is therefore dominated by large errors, MAE treats all deviations linearly. This makes MAE more robust to outliers and less sensitive to noise artifacts such as salt-and-pepper noise. Furthermore, MAE tends to preserve sharp edges and high-frequency details, making it particularly useful for images with structural discontinuities, such as beam physics and medical imaging data.

Regularization was introduced through the Total Variation (TV) loss, which penalizes abrupt intensity variations between neighboring pixels. By encouraging piecewise-smooth reconstructions, TV loss suppresses checkerboard artifacts and random noise. Ideally, TV loss decreases gradually as the network learns to smooth noise while retaining edge information; an increase may indicate excessive smoothing and loss of fine detail.

Finally, the Gradient Difference Loss (GDL) was applied to preserve sharp edges and high-frequency structures, complementing TV loss. While TV promotes smoothness, GDL emphasizes edge alignment with the target image, making it particularly effective in tasks such as super-resolution and deblurring. A consistent decrease in GDL reflects improved recovery of fine structures, whereas stagnation suggests difficulty in reconstructing high-frequency details.

The Structural Similarity Index Measure (SSIM), a perceptual metric, was also evaluated but ultimately not implemented in the final optimization framework. SSIM compares images in terms of luminance, contrast, and structural content, thereby providing an assessment more closely aligned with human visual perception than pixel-wise measures such as PSNR. Unlike PSNR, which quantifies error magnitude without accounting for perceptual factors, SSIM emphasizes structural fidelity, making it particularly sensitive to edges and fine details. Empirically, SSIM values increased during training, albeit at a slower rate than PSNR, reflecting its heightened sensitivity to structural variations. These observations highlighted that conventional quantitative metrics alone were insufficient to fully capture perceived image quality in the absence of ground-truth references. It is noteworthy that DCNNs have been observed to reproduce certain perceptual laws, reflecting similarities with mechanisms of human visual perception. However, the emergence of such perceptual properties is not guaranteed; their presence depends strongly on the characteristics of the training image and, more generally, on the learned weight distributions within the model [44]. Notwithstanding, the design of training strategies plays a critical role in determining whether DCNNs capture perceptual laws, with direct implications for their ability to preserve structural fidelity and perceptual quality in image restoration tasks according to the concept of Gestalt [45, 46].

The beam emittance area was quantified using the surface of the corresponding phase-space ellipse, serving as a task-specific metric. This surface evolves dynamically during optimization, and the calculation was terminated once the beam area reached its maximum value, prior to collapse induced by over-denoising. Initial denoising attempts did not yield significant improvements in beam area estimation for RMS emittance calculations. To overcome this limitation, the grayscale image representation was reformulated as a three-rank tensor and adopted as the new model target, after which the network was retrained accordingly during a specific alignment process.

## 6. REGULARIZATION

A regularization of DCNN models is provided by early stopping (ES) mechanisms. This technique involves stopping training as soon as the validation error starts to increase. This prevents the models from entering a phase of overfitting [47]. The associated metrics are usually defined by the generalization error, namely the difference between the error on the training data and that on the test/validation data. An increase in this error over iterations is a clear indicator of this phenomenon. In supervised learning, a validation set is used to detect overfitting and apply ES e.g. by monitoring validation loss. However, our DCNN is unsupervised and only one corrupted image is used as input, and no clean target or validation set are available. So, standard techniques like Correlation-Driven Stopping Criterion (CDSC) cannot be used directly. Adapted strategies to simulate validation behavior and introduction of heuristic features for ES as WMV and EMV are designed specifically for our DCNN model.

- ES-WMV: ES via Windowed Moving Variance. The core concept here is the monitoring of the variance between the successive DCNN outputs across all the training iterations. It is based on the key observation that a peak in variance often aligns with the optimal denoising point. The implementation manages to store the model outputs at regular intervals, compute the variance between consecutive outputs, and finally track the variance progression through the whole training process. The stopping criterion allows to end the training when the variance reaches its peak value (indicates transition from signal recovery to noise fitting). This is efficient for the cases where the image outputs can be monitored.

- ES-EMV: Expected Moving Variance (enhanced version). The improvements over previous WMV concept are a higher memory-efficiency as it doesn't require storing all outputs, only a subset of them, reducing overhead, and a lighter computation as it uses a sliding window approach. The principle is to sustain a continuously updated window of past outputs, calculate the variance as the result of the expression: (*current_output* - *window_av*g)², and update the estimates iteratively. The stopping condition is triggered when the variance estimate peaks and shows consistent downward trend afterward. This is advantageous for long training runs with high-resolution outputs.

The several ES mechanisms were evaluated and two of them were successfully incorporated in the model to prevent overfitting and optimize the denoising process: Pseudo Validation Loss (PVL) using K-Fold masks and CDSC (Correlation-Driven Stopping Criterion), and Expected Moving Variance (EMV) to monitor stability in output changes. By integrating these into the training loop, we aimed to identify the optimal stopping point for denoising each image. This significantly improved the accuracy of beam surface area estimation and hence the RMS emittance measurement. When the model is run for 10,000 iterations, the number of iterations at which ES occurs, based on hybrid criteria such as K-Fold CDSC and Expected Moving Variance (EMV), shows a strong correlation with the evolution of RMS emittance. This suggests a clear relationship between the ES iteration count and emittance behavior, indicating that the combination of the DCNN and ES is a promising approach for enhancing beam diagnostics and improving the accuracy of emittance measurements. The model parameters were systematically monitored and adjusted to achieve optimal regularization. All metrics were recorded and plotted at each iteration, see examples in Fig. 9. The points of improvement stagnation (plateau) were then identified, corresponding to the iterations where the metrics reached their optimal values. In addition, the onset of degradation was detected, for example at the minimum NIQE or maximum Tenengrad values. Consequently, the ES criterion was defined to occur slightly before or at the peak performance point.

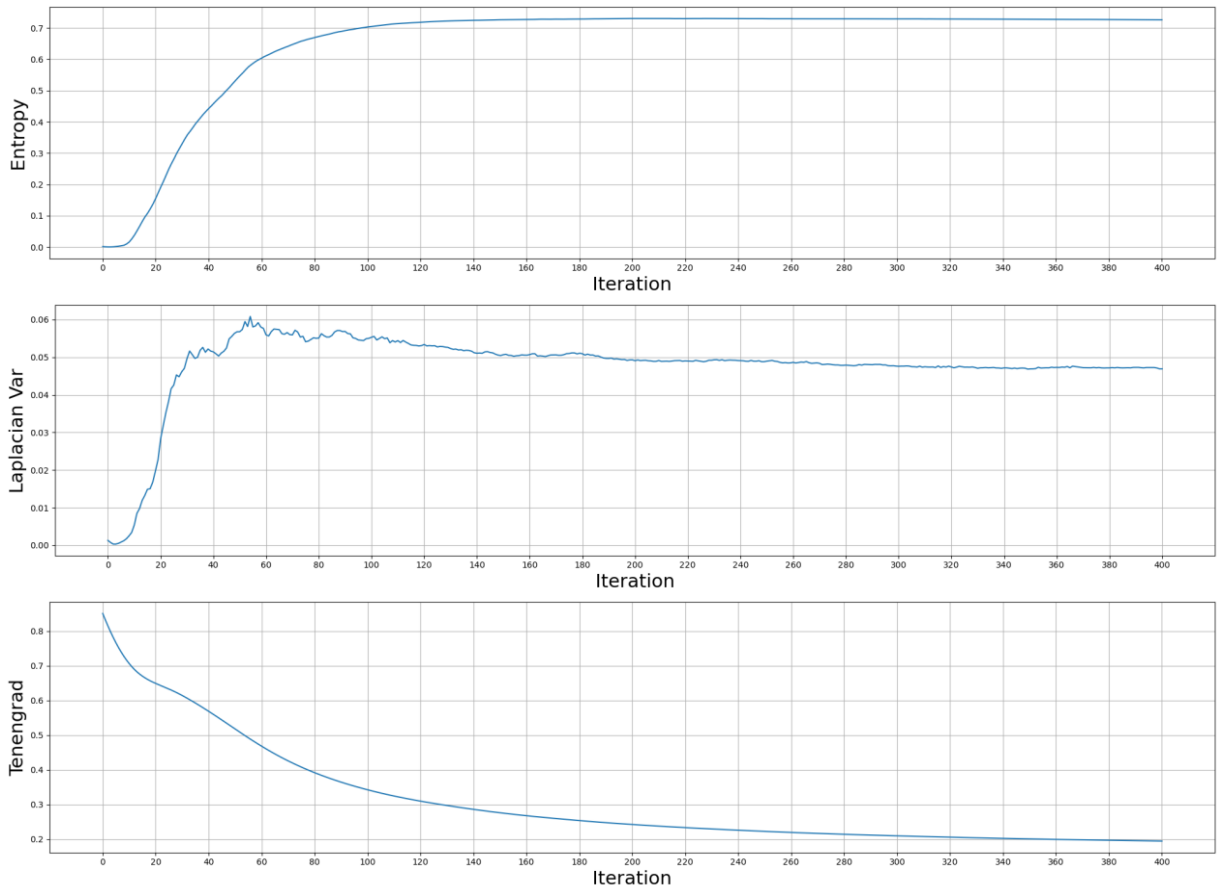

Figure 9: Shannon image entropy, Laplacian variance, and Tenengrad regularization metrics are recorded and plotted at each iteration (top down figures). The points of improvement stagnation are identified, corresponding to the iterations where the metrics achieve their optimal values.

## 7. ALIGNMENT

In order to ensure that model behave in accordance with our goals and expectations, and despite it is difficult to fully specify all behaviors of the targets in advance, a few evaluation metrics with physical meaning were used to fit the model. They are based on the beam intensity profile (1D distribution DC) and on the RMS-emittance (2D transverse phase-space distribution), see Fig. 10. Two different calculation methods were developed, namely the area of the two-dimension RMS emittance figure (so called beam_area_emittance_log function, annex B), and DBSCAN (see beam_area_dbscan_log) [40]. Using the beam distribution area in the two-dimensional phase space as the evaluation criterion (surface of RMS ellipse), we observe a stable region following a local minimum, which indicates an optimal physical value. Beyond this point, the beam area increases, corresponding to the reinjection of noise into the reconstructed image i.e. the generation of false positives. The transition between these regimes is gradual and characterized by a shallow slope. Note that we introduced an index to quantify the physical RMS emittance, the detailed assessment of which will be addressed in a future work.

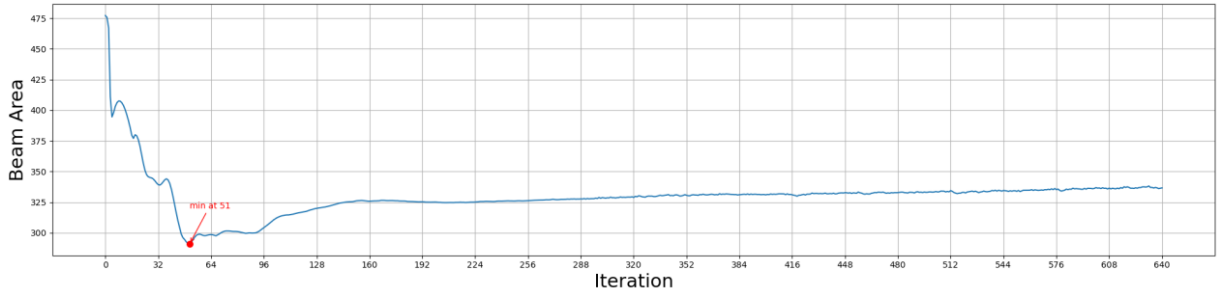

Figure 10: Area of the beam particle distribution in the two-dimensional phase space (RMS with a relative representative index) as a function of the iteration number (max 640). The local minimum is followed by a gradual increase of the area and a stable region.

We defined a dedicated procedure in order to align statistics and physics, namely, we analyzed the number of iterations at which the model terminates under ES, using a hybrid condition based on scores derived from the K-Fold CDSC and Expected Moving Variance statistical criterion. This result was then compared with the optimal iteration count estimated by tracking the evolution of the beam area over an extended run without ES up to 100,000 iterations. When the two values are in close agreement, this indicates consistency between the ES-based results and those obtained from the physical criterion of beam emittance. In cases of discrepancy, the model parameters and hyperparameters are adjusted, and the evaluation metrics refined, to achieve a balance between the two approaches.

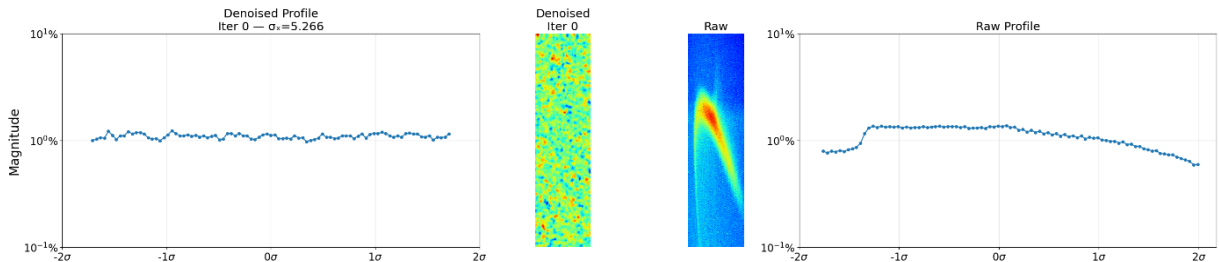

Figure 11: Profile of beam intensity distribution at different radii (σ). Left is the initial profile (iteration 0) with corresponding noise input. Right is the corrupted emittance figure with the beam profile distribution.

After 400 iterations, the emittance figure is considered as denoised, see Fig. 12. The intermediate iterations illustrating the reconstruction process from the input noise are provided in the annex D. In the best case, it is possible to determine the beam distributions up to seven standard deviations. This capability depends not only on the SNR and the noise type but also on the completeness of the available signal data.

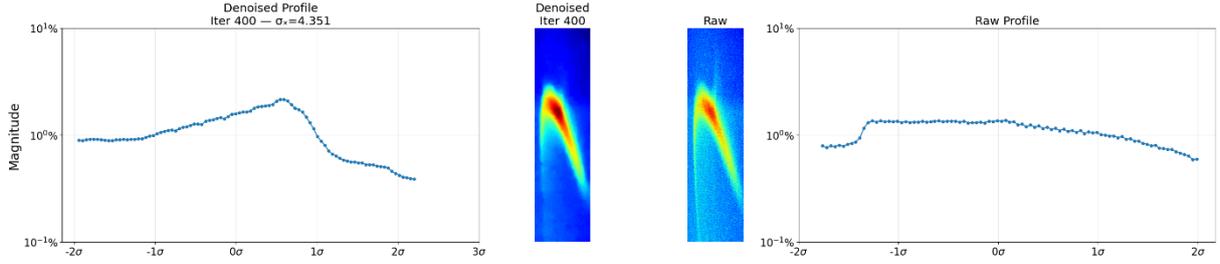

Figure 12: Cleaned profile of beam intensity distribution at different radii (σ) and denoised emittance figure after 400 iterations (left), and the corrupted emittance figure with the corresponding profile (right).

In our measurements and after image reconstruction, the effective resolution is substantially enhanced, with detectable transverse amplitudes extending beyond seven standard deviations from the beam core, see Fig. 13. At this level of sensitivity, the system is capable of resolving particle populations whose local densities fall below $10^{-4}$ of the total beam intensity, thereby capturing the outermost regions of the halo with exceptional fidelity. This extended dynamic range enables a quantitative characterization of halo structures that, to our knowledge, has not been previously achieved with this kind of emittance scanner.

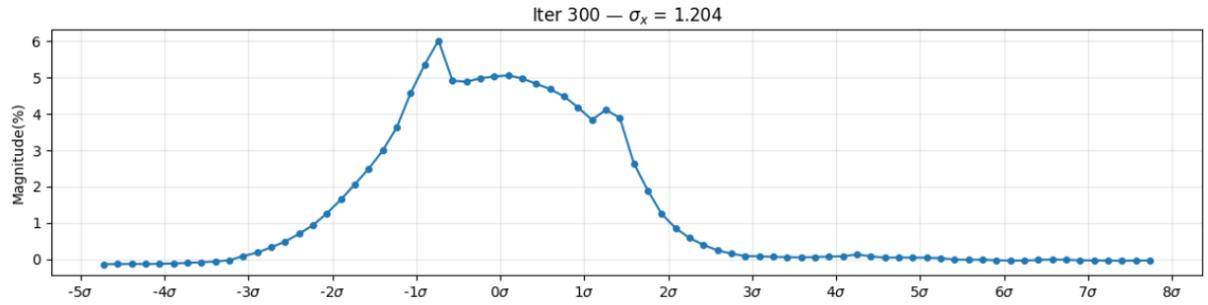

Figure 13: Typical transverse beam profile with extended radial resolution at 7 standard deviations obtained after data cleaning with DCNN model.

## 8. DISCUSSION

Our DCNN framework, based on the DIP model and new early stopping strategies, is particularly well suited for the reconstruction of natural images as well as those that characterize the beam dynamics of particle accelerators. During training, the network adjusts its weights to restore a noisy image by first capturing the underlying structure before adapting to the noise. This property enables effective denoising of the emittance figures produced with a large variety of beams and facilities without reliance on external datasets. The approach preserves fine details and local textures with high fidelity, even in the absence of GT/noisy pairs or external supervision. We observed that early stopping is critical to prevent noise reinjection, and that the model exhibits sensitivity to hyperparameters and architecture. To mitigate overfitting, we developed self-supervision techniques based on regularization and noise constraints, which proved effective in stabilizing performance. When the model is run for 10,000 iterations, the number of iterations at which early stopping occurs, based on hybrid criteria such as K-Fold CDSC and EMV, shows a strong correlation with the evolution of RMS emittance. This suggests a clear relationship between the early stopping iteration count and emittance behavior, indicating that the combination of DIP and early stopping is a promising approach for enhancing beam diagnostics and improving the accuracy of emittance measurements. Note that, scaling the framework to large image datasets introduces new computational and methodological challenges.

    Alternative self-supervised approaches, such as Noise2Void and Noise2Self, operate by masking target pixels and predicting them from the noisy context. These methods assume that noise is conditionally independent of the neighborhood used for prediction. While global training enables fast inference and stable results for additive, noise-independent data when large sets of noisy images are available, performance deteriorates in the presence of spatially correlated or signal-dependent noise, or when masking violates the independence assumption. Such limitations often result in blurred fine details and reduced structural fidelity.

    In contrast, our DCNN framework provides an elegant, data-free solution for denoising single images. It achieves strong detail preservation at low computational cost, making it a promising tool for beam diagnostics and other applications where ground-truth data and large training sets are unavailable. Furthermore, the promising

results obtained in the present study motivate the pursuit of more systematic analyses and the establishment of a comprehensive benchmarking framework. A particularly relevant direction for future work involves conducting a variational study in which beam parameters, specifically diameter, intensity, and resolution, are incrementally modified. Such controlled perturbations would yield a substantial number of image samples across different configurations. Although this procedure is inherently time-consuming, it would enable the generation of paired image sets designed to stimulate model learning through diverse representational challenges. The resulting dataset could subsequently serve as a foundation for training supervised models, which in turn would facilitate rigorous comparative evaluations.

## 9. CONCLUSION

This project demonstrates that a framework based on a combination of a DCNN model with well-designed early stopping strategies, can significantly enhance image denoising for applications related to beam diagnostics for the particle accelerator domain. Despite the lack of training datasets and reference images, our approach provides significant results through unsupervised methods and careful evaluation techniques. The reconstruction of beam emittance images, with particular emphasis on the halo extending from the beam core to the outer edges, has led to a substantial improvement in scan resolution through effective noise suppression and signal capture from the background. These results were obtained under highly unfavorable conditions, including very low signal-to-noise ratios, targets with heterogeneous and irregular shapes, non-Gaussian transverse distributions, and small, non-annotated datasets. The achievable resolution is significantly extended, with measurable amplitudes exceeding seven standard deviations in the best case, enabling a representation of the halo with a level of precision that is, to our knowledge, unprecedented with our emittance scanner. Notably, this approach enabled the detection of a halo previously unobserved at one of the pilot installations.

In addition to its scientific contributions, the proposed solution is characterized by its frugality: it requires minimal computational resources, allows computing entirely in CPU mode, and does not rely on cloud computing, data centers, or internet connectivity. This lightweight design results in a markedly reduced carbon footprint, aligning the methodology with sustainable development objectives.

## ACKNOWLEDGEMENTS

We gratefully acknowledge the CNRS Nuclei & Particles laboratories that contributed to this work and provided access to the beams, Dr T. Durand for his collaborative spirit and insightful discussions over the years, as well as the Fidle@CNRS team who allowed us to get our foot in the door.

Code and additional material are available at: https://github.com/FrancisO67/Beam-Image-Restoration

Credit authorship contribution statement
F.R. Osswald: Supervision, Methodology, Validation, Writing – original draft, review & editing.
M. Chahbaoui: Software, Coding, Model Optimization, Tests, Writing – original draft.
X. Liang: Software, Coding, Model Optimization, Tests, Writing – original draft.

Declaration of competing interest
The authors declare that they have no known competing financial interests or personal relationships that could have appeared to influence the work reported in this paper.

# ANNEX A:
# LIBRARIES

```python
import os, sys
import pandas as pd
import numpy as np
import matplotlib.pyplot as plt
import torch
from torch.optim import Adam
import torch.nn.functional as F
import math
from piq import brisque
import scipy
from pypiqe import piqe
from skimage.measure import shannon_entropy
from scipy.stats import pearsonr
from sklearn.model_selection import KFold
```

# ANNEX B:
# METRICS

```python
metrics_dict = {
"MSE Loss": mse_loss_log,
"MAE Loss": mae_loss_log,
"TV Loss": tv_loss_log,
"GDL Loss": gdl_loss_log,
"Total Loss": total_loss_log,
"NIQE": niqe_log,
"BRISQUE": brisque_log,
"Entropy": entropy_log,
"Laplacian Var": laplacian_log,
"Tenengrad": tenengrad_log,
"Beam Area (Emittance)": beam_area_emittance_log,
"Pseudo Validation Loss": pseudo_val_loss_log
}
plot_metrics(file_signature, metrics_dict, station_name, image_name, metrics_out_dir, metric_to_plot=None, SAVE=True)
```

# ANNEX C:
# OPTIMIZATION

```python
# Compute loss
(mse_val, mae_val, grad_val, tv_val), total_loss = Loss_function(
out,
w_mse=loss_config["w_mse"],
w_mae=loss_config["w_mae"],
w_tv=loss_config["w_tv"],
w_gd=loss_config["w_gd"],
mask=mask # only training pixels included
)
total_loss.backward()
# Update the model parameters based on the gradients we've just computed during total_loss.backward().
optimizer.step()
```

## ANNEX D:
## EMITTANCE FIGURES

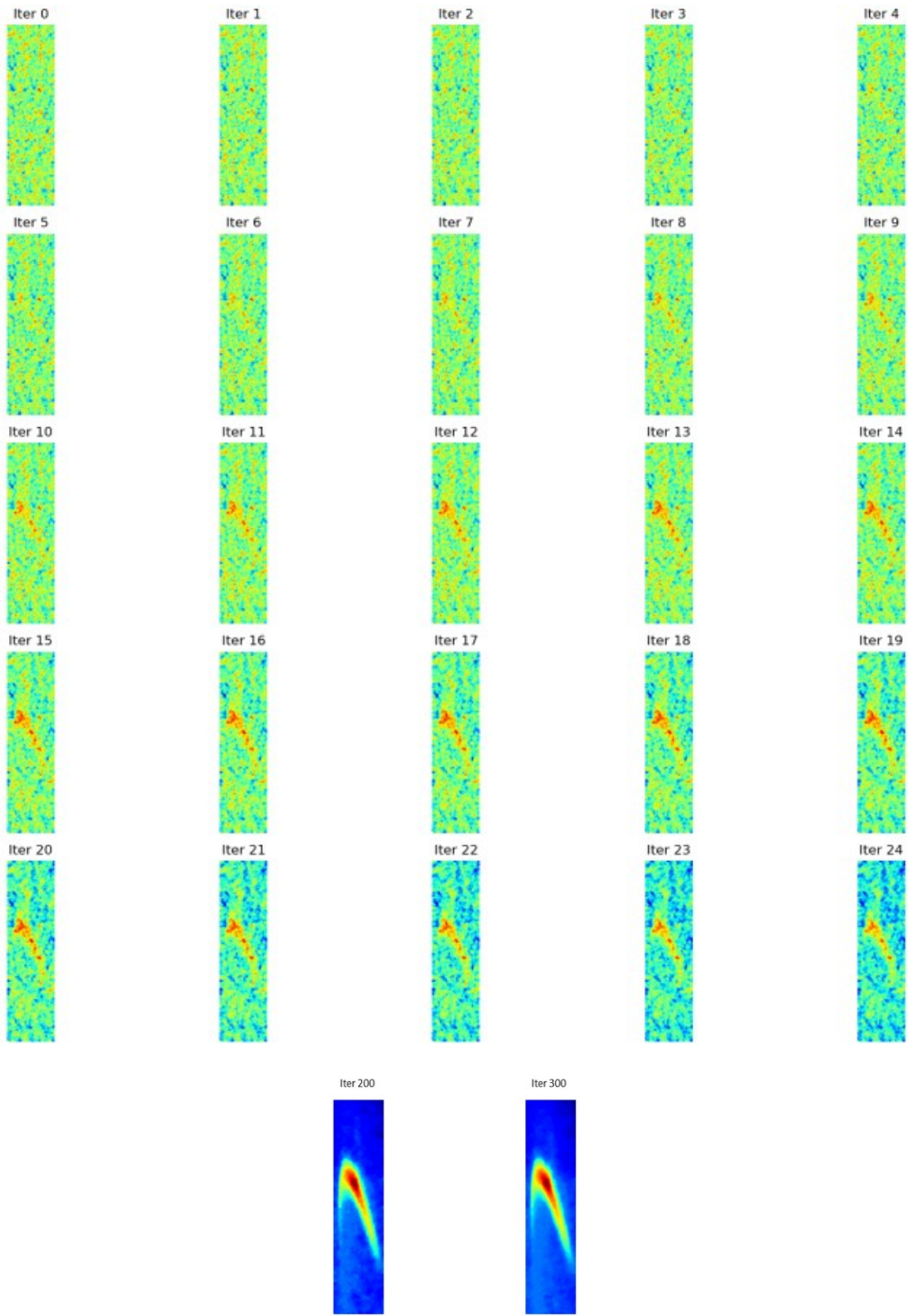

Emittance figures for different iterations (0-24) showing the reconstruction process starting from the input noise (Iter 0). After 300 iterations, the emittance figure is considered as denoised (bottom).